\documentclass[review]{elsarticle}

\usepackage{lineno,hyperref}
\modulolinenumbers[5]
\usepackage{algorithm}
\usepackage{amsmath}  
\usepackage{algorithmic}
\usepackage{amsfonts}
\usepackage{bm}
\usepackage{dsfont}
\usepackage{booktabs}

\begin{document}

\begin{frontmatter}

\title{Gaussian Processes for Missing Value Imputation}


\author[inibica]{Bahram Jafrasteh\corref{mycorrespondingauthor}}
\ead{jafrasteh.bahram@inibica.es}

\author[uam]{Daniel Hernández-Lobato}
\ead{daniel.hernandez@uam.es}
\author[inibica,hpum]{Simón Pedro Lubián-López}
\ead{simonp.lubian.sspa@juntadeandalucia.es}
\author[inibica,hpum,uca]{Isabel Benavente-Fernández}
\ead{isabel.benavente@uca.es}

\cortext[mycorrespondingauthor]{Corresponding author}

\address[inibica]{Biomedical Research and Innovation Institute of Cádiz (INiBICA) Research Unit, Puerta del Mar University, Cádiz, Spain}
\address[uam]{Computer Science Department, Universidad Autónoma de Madrid, Madrid, Spain}
\address[hpum]{Division of Neonatology, Department of Paediatrics, Puerta del Mar University Hospital, Cádiz, Spain}
\address[uca]{Area of Paediatrics, Department of Child and Mother Health and Radiology, Medical School, University of Cádiz, Cádiz, Spain}

\begin{abstract}
Missing values are common in many real-life datasets.
However, most of the current machine learning methods 
can not handle missing values. This means that they 
should be imputed beforehand. Gaussian Processes (GPs) are 
non-parametric models with accurate uncertainty estimates
that combined with sparse approximations and stochastic 
variational inference scale to large data sets.
Sparse GPs can be used to compute a predictive 
distribution for missing data.  
Here, we present a hierarchical composition 
of sparse GPs that is used to predict missing values at
each dimension using all the variables from the other dimensions.
We call the approach missing GP (MGP). MGP can be trained 
simultaneously to impute all observed missing values. Specifically, 
it outputs a predictive distribution for each missing value that
is then used in the imputation of other missing values.
We evaluate MGP in one private clinical data set and four 
UCI datasets with a different percentage of missing values.
We compare the performance of MGP with other state-of-the-art methods 
for imputing missing values, including variants based on sparse GPs 
and deep GPs. The results obtained show a significantly better 
performance of MGP.
\end{abstract}


\end{frontmatter}

\section{Introduction}\label{Intro}

Data in real-life sciences are often noisy, stored in databases, and 
contain missing values. This is particularly the case of clinical data. 
Specifically, our work is motivated by a clinical data set of newly born premature 
infants. This dataset has different variables that are related to various 
diagnoses at the current and previous states of life and contains, as measurements, 
for example, the total brain volume estimated using ultrasound images. 
Often, for each infant, it is not possible to measure and store all variables 
considered in the dataset. Therefore, there are several missing values associated 
to some instances. Having missing values is also very common in datasets from other domains. 
Some examples include those related to lab measurements or shopping platforms 
\citep{Missingdata01}. 

Machine learning models can learn the underlying data structure, 
capture high dimensional and complex relationships between different variables, and hence, 
estimate the true distribution of the data to make predictions on previously 
unseen data \citep{bishop2006}. They allow us to make a better decision, create high-quality clusters, 
detect outliers in the data, and make more accurate predictions of a parameter of interest.
However, machine learning methods for regression, classification and clustering methods 
that ignore the impact of the missing data can lead to sub-optimal models without good 
enough generalization performance. Therefore, taking into account missing values and not 
just ignoring data instances with missing values is a critical step in a machine learning method.

To be able to take into account missing values when fitting a machine learning method it is
important to know what type of missing values one can find in practical applications.
In particular, there are three kinds of missing value 
mechanisms described in the literature:
\begin{itemize}
    \item Missing completely at random (MCAR): The missingness mechanism is 
not related to any observed value or unobserved values from the dataset.
Therefore missing values appear completely at random on the observed instances.
    \item Missing at random (MAR): Having missing values for one variable is related to the
value of some other observed variable or variables from the dataset. For example, men are 
more likely to tell their weight and women are less likely. Therefore, the missing value mechanism 
for weight is explained by gender.
    \item Missing not at random (MNAR): The missing instances are related to some values of 
the same attribute. For example, if an attribute considers the level of education, people
may be ashamed of answering that they have the lowest education level and they may not fill 
that information.
\end{itemize}

It is common to assume the missing mechanism as MAR and impute missing values 
using traditional methods or machine learning algorithms. The imputation of missing values 
plays a key role in the final performance since the chosen algorithm directly impacts the final
model. Removing instances with missing values from a dataset and training a model with all remaining 
data is considered a minimal and simple approach that is expected to result in a suboptimal
performance. Moreover such a model will not be able to consider new instances for prediction 
with missing values.  

Most machine learning methods for regression, classification, and/or clustering 
inherently can not deal with missing values. Hence, it is needed to provide a 
way to impute this missing data and/or change the machine learning method. 
The simplest approach is to impute the data with their mean/median values across 
data instances. However, several studies show that these approaches are not sufficient. 
The generated model can still be sub-optimal \citep{beaulieu2017missing, ryu2020denoising}. 
More sophisticated methods should be used to find the true distribution of missing values and 
impute them.

Recovering latent values associated to missing values can help the final decision makers 
to improve their predictions. Moreover, it can be useful to better understand the dependence 
of the target variable to predict on the explaining attributes. The data distribution of 
missing values can be extracted using a predictor for the missing value and the 
corresponding associated prediction uncertainty. This prediction uncertainty 
will be able to capture what are the potential values that a missing observation may 
have had. This uncertain is expected to be important, since there is evidence that 
incorporating input noise in a final model can improve overall prediction performance 
\citep{villacampa2020multi}.

One of the well-known non-parametric machine learning approaches 
with a probabilistic nature is a Gaussian process (GP) \citep{williams2006gaussian}.
A GP can output a predictive distribution for the target variable that takes into 
account prediction uncertainty. This uncertainty arises from intrinsic noise in the data and 
also because of the fact that we have a finite amount of training data. Given $N$ 
observation points, the training of a GP requires an inversion of $N \times N$ covariance 
matrix, which is very expensive for a large $N$. Therefore, as 
the number of training instances increases, approximate techniques will be used for 
the computation of the predictive distribution.

One of the most popular approximations methods to deal with the scalability of GPs 
is based on sparse inducing points representations \citep{titsias2009variational, snelson2006sparse}. 
In the sparse variational Gaussian process (SVGP) $M \ll N$ inducing 
points are optimized alongside with other hyper-parameters using variational inference 
\citep{hensman2015mcmc, villacampa2017scalable}. GPs can scale to a very large datasets using 
a combination of sparse approximations and stochastic optimization techniques \citep{hensman2015scalable}.

A concatenation of GPs corresponds to a deep GP (DGP). DGPs have been proposed to improve the performance 
of single-layer GPs, similarly to what happened in the case of multilayer neural 
networks \cite{damianou2013deep, bui2016deep,salimbeni2017doubly}. DGPs overcome 
some limitations of the single layer sparse GPs such as the expressiveness of kernel/covariance function 
while still being able to output a predictive distribution and scale to large datasets \cite{salimbeni2017doubly}.
DGPs and SVGPs can be used with multiple input and multiple outputs to learn the latent 
representation of the data and recover the data distribution. However, DGPs do not consider 
sequential relations between a set of variables in the dataset. Recurrent GPs have been introduced 
in \cite{mattos2016recurrent} for sequential data sets.  

In this work we are inspired by the DGP architecture \cite{salimbeni2017doubly} and the 
recurrent GP to develop a new method of imputing missing values. The method is a hierarchical 
composition of GPs, where there is a GP per dimension that predicts the missing values for 
that dimension using all the variables from the other dimensions. Of course, for this to
work, an ordering on the dimensions has to be specified and also an initial value for the
missing values. 

Our method starts with the dimension that has the largest standard deviation (before standardizing the data).
The missing values of that dimension are predicted using a GP that receives as an input all the 
other dimensions. Missing values in those dimensions are simply estimated initially using the mean 
value across the corresponding dimension. After this, a second GP is used to predict the missing values
of the dimension with the second largest standard deviation (before standardizing the data). This second GP
also receives all the other dimensions as an input. Importantly, however, this second GP
receives as an input for the missing values corresponding to the dimension with the largest 
number of missing values the predictions given by the first GP. This process is then repeated 
iteratively for the total number of dimensions with missing values, using the predictions of the 
previous GPs for the attributes with missing values. 

Given the imputed missing values by the process described and the observed data, we then have a 
last GP that predicts the target variable. That last GP receives as an input the observed data 
and the predictions given by the GPs in charge of imputing the missing values. Therefore, with 
the process described all the missing values have an associated predictive distribution which is 
taken into account by the last GP for prediction. All the GPs are trained at the same time. 

We have validated the method described, called missing GP (MGP), using one private clinical data and 
four datasets extracted from UCI repository \citep{Dua2019}. The private data is provided by the
"perinatal brain damage" group at the Biomedical Research and Innovation Institute of Cádiz (INiBICA) and the Puerta del Mar University Hospital, Cádiz, Spain.

The contributions of this work are:
\begin{itemize}
    \item A new method based on concatenated GPs is introduced for imputing missing values.
    \item The method outputs a predictive distribution for each missing value in the dataset. This can be used for missing value imputation.
    \item The final model can be trained simultaneously and can be scaled to large data sets.
\end{itemize}
The manuscript is organized as follows: In the next section, we briefly describe 
the Gaussian processes, DGPs, and then we explain the proposed method. The configuration 
of the experiments and the datasets are explained in Section \ref{Exp}. 
In Section \ref{Results_discussion}, we discuss the obtained results and, finally, 
Section \ref{sec:conclusion} presents the conclusions.

\section{Gaussian Processes for Missing Data}

This section gives a briefly introduction to Gaussian process (GPs) and Deep GPs (DGPs). It provides
the needed background to correctly explain the proposed method for learning under missing values
using GPs. We call such a method missing GP (MGP). 

\subsection{Gaussian Processes}

A Gaussian process (GP) is a stochastic process 
whose values at any finite set of points follow a multi-variate Gaussian distribution \citep{rasmussen2005book}. 
From a machine learning perspective, a GP is used as a prior over a latent function $f$, where the posterior of 
that function computed using the observed data is another GP. This results in a non-parametric 
machine learning method whose level of expressiveness grows with the dataset size.  Consider a set of points 
$\mathcal{D}=\{(\mathbf{x}_i, y_i)\}_{i = 1}^N$ and $y_i = f(\mathbf{x}_i) + \epsilon_i$, where $\epsilon_i$ is 
a Gaussian noise with variance $\sigma^{2}$. A GP prior for $f$ is typically specified by 
a mean function $m(\mathbf{x})$ and a covariance function $k(\mathbf{x}, \mathbf{x}')$ with a 
trainable parameter $\bm{\theta}$. Assuming a zero mean function, given a dataset 
$\mathcal{D}$, the predictive distribution for the value of $f$, $f^\star$, at a new test point $\mathbf{x}^\star$ is Gaussian.
Namely,
\begin{align}
p(f^\star |\mathcal{D}) = \mathcal{N} (\mu(\mathbf{x}^\star), \mathbf{\sigma}^{2}(\mathbf{x}^\star))\,,
\end{align}
with mean and variance given by
\begin{align}
\mu(\mathbf{x}^\star) &= \mathbf{k}(\mathbf{x}^\star)^\text{T} 
(\mathbf{K}+\mathbf{\sigma^{2} \mathbf{I}})^{-1} \mathbf{y}\,, \\
\sigma^{2}(\mathbf{x}^\star) &= 
k(\mathbf{x}^\star,\mathbf{x}^\star) - \mathbf{k}(\mathbf{x}^\star)^\text{T} (\mathbf{K}+\mathbf{\sigma^{2} \mathbf{I}})^{-1} 
\mathbf{k}(\mathbf{x}^\star)\,,
\label{Eq.GPvar}
\end{align}
where $\mu(\mathbf{x^\star})$ and $\sigma^{2}(\mathbf{x}^\star)$ denotes the predictive
mean and variance, respectively. $\mathbf{k}(\mathbf{x}^\star)$ is a vector with  the covariances 
between $f(\mathbf{x}^\star)$ and each $f(\mathbf{x}_i)$, simply given by $k(\mathbf{x}^\star,\mathbf{x}_i)$,
with $k(\cdot,\cdot)$ the covariance function.  $\mathbf{K}$ is a $N \times N$ matrix with the covariances 
between each $f(\mathbf{x}_i)$ and $f(\mathbf{x}_j)$ in the training set. That is, $K_{ij} = k(\mathbf{x}_i,\mathbf{x}_j)$. 
$\mathbf{I}$ stands for the identity matrix. 

The learning of the hyper-parameters $\bm{\theta}$ can be
done by maximizing the marginal likelihood of the model. Namely, $p(\mathbf{y}|\bm{\theta})$, 
which is Gaussian \citep{rasmussen2005book}. It is possible to show that the marginal likelihood penalizes models
that either too simple or too complicated to explain the observed data \citep{bishop2006}.

Importantly, GPs are unsuitable for large datasets as they need the inversion of matrix $\mathbf{K}$, with 
a computational complexity in $O(N^{3})$. However, one can use sparse GPs to overcome this problem. 
Sparse GPs are explained in the next section.

\subsection{Sparse Gaussian Processes} \label{SVGP}

One can reduce the computational cost of GPs with introducing a $M \ll N$ additional data 
called inducing points $\mathbf{Z}=(\mathbf{z}_1,\ldots,\mathbf{z}_M)^{T}$ \cite{snelson2006sparse}. 
The inducing points are in the same space as each $\mathbf{x}_i$.
Let $\mathbf{u}=(f(\mathbf{z}_1),\ldots,f(\mathbf{z}_M))^\text{T}$ be the process 
values at the inducing points  and $\mathbf{f}=(f(\mathbf{x}_1),\ldots,f(\mathbf{x}_N))^\text{T}$ 
be the process values at the training data. As a consequence of using a GP with mean function $m(\cdot)$, 
$p(\mathbf{u}) \sim \mathcal{N}(\mathbf{m}(\mathbf{Z}), \mathbf{K}(\mathbf{Z}, \mathbf{Z}))$, 
with $\mathbf{K}(\mathbf{Z}, \mathbf{Z})$ the covariance matrix that results from evaluating the 
covariance function $k(\cdot,\cdot)$ on the inducing points and $\mathbf{m}(\mathbf{Z})$ the vector
of the prior GP means for each $f(\mathbf{z}_j)$. 

Consider now the use of variational inference (VI) to find an approximate posterior for 
$\mathbf{f}$ and $\mathbf{u}$ given the observed data \citep{titsias2009variational}. Specifically, 
the goal is to find an approximate posterior $q(\mathbf{f}, \mathbf{u})$ which resembles to true posterior 
$p(\mathbf{f}, \mathbf{u}| \mathbf{y})$. Following, \cite{hensman2013gaussian,titsias2009variational} 
we can specify a constrained form for $q$. Namely,
\begin{align}
q(\mathbf{f}, \mathbf{u})=p(\mathbf{f}|\mathbf{u})q(\mathbf{u})\,,
\end{align}
where the first factor is fixed and given by the GP predictive distribution, and 
the second factor is a tunable multi-variate Gaussian $q(\mathbf{u})=\mathcal{N}(\mathbf{r}, \mathbf{S})$. 

One can marginalize out $\mathbf{u}$ in order to compute the mean and variances of the 
predictive distribution at the inputs. That is, $q(f(\mathbf{x}_i))$ is Gaussian with parameters
\begin{align}
\mu_{\mathbf{r}, \mathbf{Z}}(\mathbf{x}_i) &= m(\mathbf{x}_i) + \mathbf{\alpha}(\mathbf{x}_i)^{T}
(\mathbf{r} - m(\mathbf{Z}) ) \label{Eq.SGPvar0}\,, \\
\sigma^2_{\mathbf{S}, \mathbf{Z}}(\mathbf{x}_i, \mathbf{x}_i) &= 
k(\mathbf{x}_i, \mathbf{x}_i) -  \mathbf{\alpha}(\mathbf{x}_i)^{T} 
(\mathbf{K}(\mathbf{Z},\mathbf{Z}) - \mathbf{S})  \mathbf{\alpha}(\mathbf{x}_i) \,,
\label{Eq.SGPvar}
\end{align}
where $\mathbf{\alpha}(\mathbf{x}_i)= \mathbf{K}(\mathbf{Z},\mathbf{Z})^{-1} \mathbf{k}(\mathbf{Z},\mathbf{x}_i)$,
with $\mathbf{k}(\mathbf{Z},\mathbf{x}_i)$ the vector that results from evaluating 
$k(\mathbf{z}_j,\mathbf{x}_i)$ for $j=1,\ldots,M$. 

The variational parameters $\mathbf{Z}, \mathbf{r}, \mathbf{S}$ and hyper-parameters are 
optimized by maximizing the evidence lower bound $\mathcal{L}$ (ELBO) on the log-marginal 
likelihood, as described in \cite{titsias2009variational,hensman2013gaussian}. 
This is known as the sparse-variational GP  model (SVGP).
Namely,
\begin{align}
\label{Eq_elbo_SVGP}
\mathcal{L} = \mathds{E}_{q(\mathbf{f}, \mathbf{u})}[\log \frac{p(\mathbf{y},\mathbf{f},\mathbf{u})}{q(\mathbf{f},\mathbf{u})} ]\,,
\end{align}
where $p(\mathbf{y},\mathbf{f},\mathbf{u}) = \prod_{i=1}^N p(y_i|f_i) p(\mathbf{f}|\mathbf{u})p(\mathbf{u})$.
In this last expression, the first factors correspond to the likelihood and the other two factors represent 
the GP prior on $\mathbf{f}$ and $\mathbf{u}$. After some simplifications, the 
lower bound is computed as follows
\begin{align}
\mathcal{L} = \textstyle \sum_{i=1}^{N}
	\mathds{E}_{q(f_i)}[\log p(y_i|f_i)] - 
	\text{KL}[q(\mathbf{u})|p(\mathbf{u})]\,,
\label{eq:elbo_SVGP}
\end{align}
where $\text{KL}$ stands for the Kullback-Leibler divergence between the 
distributions $q(\mathbf{u})$ and $p(\mathbf{u})$, and $f_i=f(\mathbf{x}_i)$. 
Since both distributions are Gaussian, we can analytically compute the KL value. 
In the case of regression, where a Gaussian likelihood is used, the expectation 
has a closed form and there is no need to use extra approximation methods. 
Critically, the objective function, \emph{i.e.}, the lower-bound $\mathcal{L}$, 
involves a sum over training instances and hence, can be combined with mini-batch 
sampling and stochastic optimization techniques for inference on large datasets 
\cite{hensman2013gaussian}. 

Instead of a one dimensional output $y_i \in \mathbb{R}$, one can consider $D$-dimensional 
outputs. Namely, $\mathbf{y}_i \in \mathbb{R}^{D}$. These problems can be addressed by considering
$D$ independent GPs. The GP prior is changed to a factorizing prior 
of $D$ GPs. Namely,
$\prod_{d=1}^D p(\mathbf{f}^d|\mathbf{u}^d)p(\mathbf{u}^d) = p(\mathbf{F}|\mathbf{U})p(\mathbf{U})$, 
with $\mathbf{f}^d$ and $\mathbf{u}^d$ being the $d$ sparse GP process values at the training points and the 
inducing points, respectively. 
Therefore, $\mathbf{F}=(\mathbf{f}^1,\ldots,\mathbf{f}^D)$ and $\mathbf{U}=(\mathbf{u}^1,\ldots,\mathbf{u}^D)$.
Moreover, we can assume that the inducing points $\mathbf{Z}$ 
are shared across each of the $D$ different sparse GPs. The joint distribution of all the 
variables can be rewritten
\begin{equation}
p(\mathbf{Y},\mathbf{F},\mathbf{U}) = \prod_{i=1}^N p(\mathbf{y}_i|\mathbf{f}_i) 
p(\mathbf{F}|\mathbf{U})p(\mathbf{U})\,,
\label{Eq.jointdist_SVGP_y}
\end{equation}
where $\mathbf{f}_i$ is the $i$-th row of $\mathbf{F}$, a $D$ dimensional vector 
with the values of each of the $D$ latent functions at $\mathbf{x}_i$.
One can also consider a similar approximate distribution $q$. Namely, 
$q(\mathbf{F},\mathbf{U})=p(\mathbf{F}|\mathbf{U})p(\mathbf{U})$.
Then, the ELBO is 
\begin{align}
\mathcal{L} & = \sum_{i=1}^{N}\mathds{E}_{q(\mathbf{f}_i)}[\log p(\mathbf{y}_i|\mathbf{f}_i)] - \text{KL}[q(\mathbf{U})|p(\mathbf{U})]
\nonumber \\
& = \sum_{i=1}^{N}\mathds{E}_{q(\mathbf{f}_i)}[\log p(\mathbf{y}_i|\mathbf{f}_i)] - \sum_{d=1}^D 
	\text{KL}[q(\mathbf{u}^d)|p(\mathbf{u}^d)]\,.
\label{eq:elbo_SVGP_y}
\end{align}

Note that the method described can be used to map all input attributes to themselves for missing value imputation.
In this case, $\mathbf{Y}=\mathbf{X}$. Of course, for this to work one needs to have an initial guess for the
missing values so that they can be considered as input attributes of each latent function.
Missing values can be initially estimated using a simple technique such as mean imputation. 
After learning the latent representation of the inputs, the missing values can be predicted using the 
predictive distribution of the method described.

\subsection{Deep Gaussian Processes}\label{DGP}

\begin{figure}
  \includegraphics[width=\linewidth]{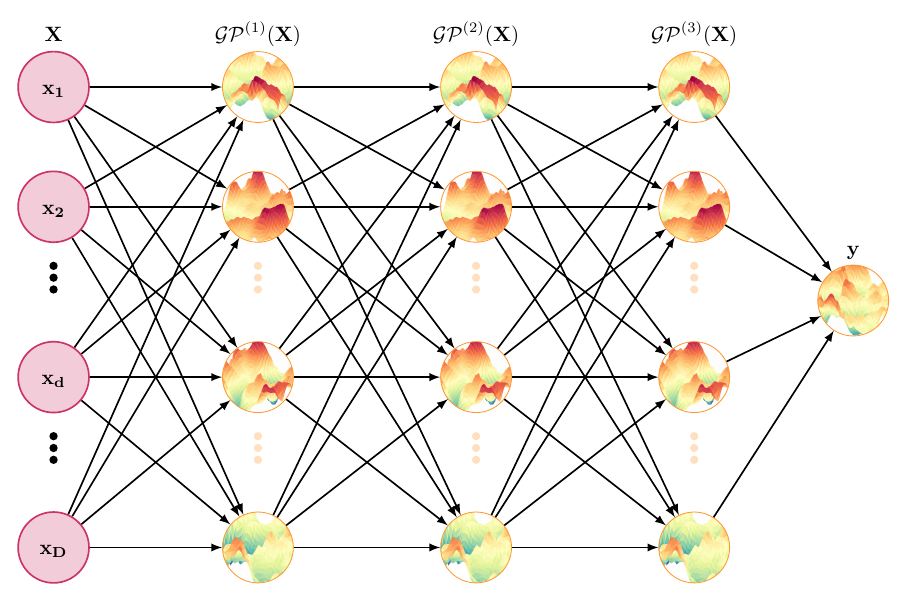}
  \caption{Three layered Deep Gaussian process.}
  \label{fig_dgp}
\end{figure}

Deep Gaussian process (DGP) \cite{damianou2013deep, bui2016deep} are a concatenation of independent GPs. Namely,
the GPs at layer $l$ receive as an input the output of the GPs at layer $l-1$, in the same spirit as a deep
neural network, but where each unit in a hidden layer is a GP. Consider a DGP of $L$ layers with $H$ units 
or GPs in each layer. Figure \ref{fig_dgp} illustrates this architecture.
Let $\mathbf{F}^{(l)}$ be the function values associated to the input points
in layer $l$. 
That is, $\mathbf{F}^{(l)}$ is a matrix of size $N \times H$.
For computational reasons, sparse GPs based on inducing points are used instead of standard GPs
in each layer. Thus, each layer $l$ has inducing points $\mathbf{Z}^{(l)}$, a noisy inputs 
$\mathbf{F}^{(l-1)}$ received from the previous layer. Note that here we assume shared inducing points for the GPs
of each layer. The inducing points values of layer $l$ are denoted by $\mathbf{U}^{(l)}$, a $M \times H$ matrix. 
Given a DGP, the joint distribution of all the variables in the model is 
\begin{align}
p(\mathbf{y}, \{\mathbf{F}^{(l)},\mathbf{U}^{(l)}\}_{l=1}^L) & = 
\prod_{l=1}^L p(\mathbf{F}^{(l)}|\mathbf{U}^{(l)}, \mathbf{F}^{(l-1)}, \mathbf{Z}^{(l)}) 
	p(\mathbf{U}^{(l)}| \mathbf{Z}^{(l)}) \times \nonumber \\
	& \quad \times \prod_{i=1}^{N} p(y_{i}|f_{i}^L)
\label{Eq_dgp_p}
\end{align}
where the inputs to the first layer are the observed data instances $\mathbf{X}$ and $f_i^L$ is 
the corresponding function value associated to the DGP at the last layer for instance $\mathbf{x}_i$.
Moreover, $p(\mathbf{F}^{(l)}|\mathbf{U}^{(l)}, \mathbf{F}^{(l-1)}, \mathbf{Z}^{(l)})$
is given by each GP predictive distribution at layer $l$, as in the single-layer sparse GP described
in the previous section. Since exact inference in the model is not tractable, approximate inference has to be used.
The work in \cite{salimbeni2017doubly} introduces a method based on variational inference
and the following form of the posterior approximation
\begin{equation}
\label{Eq_dgp_q}
\begin{split}
q(\{\mathbf{F}^{(l)}, \mathbf{U}^{(l)}\}_{l=1}^{L}) = & \prod_{l=1}^L p(\mathbf{F}^{(l)}|\mathbf{U}^{(l)}, \mathbf{F}^{(l-1)}, \mathbf{Z}^{(l)}) q(\mathbf{U}^{(l)}) \,,
\end{split}
\end{equation}
where $p(\mathbf{F}^{(l)}|\mathbf{U}^{(l)}, \mathbf{F}^{(l-1)}, \mathbf{Z}^{(l)})$ and $q(\mathbf{U}^{(l)})$ 
factorize across units in a layer as in the single-layer sparse GP described in the previous section. 
Moreover, $p(\mathbf{F}^{(l)}|\mathbf{U}^{(l)}, \mathbf{F}^{(l-1)}, \mathbf{Z}^{(l)})$
is fixed and given by each GP predictive distribution 
and $q(\mathbf{U}^{(l)})$ is a product of multi-variate Gaussian distributions that can be adjusted.
After marginalizing out $\mathbf{U}$ from each layer, the posterior is a product of Gaussian distributions
\begin{equation}
\label{Eq_dgp_qn}
\begin{split}
q(\{\mathbf{F}^{(l)}\}_{l=1}^{L}) &= \prod_{l=1}^L 
	q(\mathbf{F}^{(l)}|\mathbf{r}^{(l)}, \mathbf{S}^{(l)}, \mathbf{F}^{(l-1)}, \mathbf{Z}^{(l)}) \\&=
\prod_{l=1}^L \mathcal{N}(\mathbf{F}^{(l)}|\bm{\mu}^{(l)}, \mathbf{\Sigma}^{(l)}).
\end{split}
\end{equation}
where the mean and variance of each marginal Gaussian distribution 
are computed as in \eqref{Eq.SGPvar0} and \eqref{Eq.SGPvar}. 
For each sample $i$ and unit $h$ at layer ${l}$ the mean is 
$\mu^{(l)}_{i,h}=\mu_{\mathbf{r}_h^{(l)}, \mathbf{Z}^{(l)}}(\mathbf{f}_i^{(l-1)})$ and 
the variance  is 
$(\Sigma)_{i,h}^{(l)}=\sigma^2_{\mathbf{S}_h^{(l)}, \mathbf{Z}^{(l)}}(\mathbf{f}_i^{(l-1)},\mathbf{f}_i^{(l-1)})$, 
where $\mathbf{f}^{(l-1)}_{i}$ is the $i$-th row of $\mathbf{F}^{(l-1)}$. 
Having two distributions $p$ from \eqref{Eq_dgp_p} and $q$ from 
\eqref{Eq_dgp_q} and putting them into \eqref{Eq_elbo_SVGP}, the 
ELBO of a DGP is 
\begin{align}
\begin{split}
\mathcal{L}_{DGP} &= \textstyle \sum_{i=1}^{N}\mathds{E}_{q}[\log p(y_i|f_i^{L})] \\ & -
\sum_{l=1}^{L} \text{KL}[q(\mathbf{U}^{(l)})|p(\mathbf{U}^{(l)}| \mathbf{Z}^{(l)})]\,,
\end{split}
\label{Eq_elbo_dgp}
\end{align}
where $\mathbf{f}_i^L$ are the latent functions of the last layer associated to $\mathbf{x}_i$.
Critically, $\mathds{E}_{q}[\log p(y_i|f_i^{L})]$ is intractable and requires a Monte
Carlo approximation. This approximation can be combined with stochastic optimization techniques
for training the model \cite{salimbeni2017doubly}.

In a DGP, the predictive distributions of layer $l$ for the output associated to $\mathbf{x}_i$, 
denoted with $\mathbf{f}_i^l$, depends on the output of the previous layer $\mathbf{f}^{(l-1)}_i$. 
Let $f_{i,h}^l$ be the output of unit $h$ at layer $l$ for the data instance $\mathbf{x}_i$.
Using this property, one can use the reparameterization 
trick \citep{rezende2014stochastic, kingma2015variational} to recursively sample
$\hat{f}_{i,h}^{(l)} \sim q(f_{i,h}^{(l)}|\mathbf{r}_h^{(l)}, \mathbf{S}_h^{(l)}, 
\hat{\mathbf{f}}_{i}^{(l-1)}, \mathbf{Z}^{(l)} )$ with
\begin{equation}
    \hat{f}_{i,h}^{(l)} = \mu_{\mathbf{r}_{(l)}^h, \mathbf{Z}_{(l)}}(\hat{\mathbf{f}}^{(l-1)}_{i}) + 
	\epsilon_{i,h}^{(l)} \sqrt{\sigma^2_{\mathbf{S}_h^{(l)},\mathbf{Z}^{(l)}}(\hat{\mathbf{f}}^{(l-1)}_{i},
	\hat{\mathbf{f}}^{(l-1)}_{i})}
    \label{Eq_montcarlo}
\end{equation}
where $\mathbf{f}_{i}^{(0)}=\mathbf{x}_i$ and $\epsilon_{i,h}^{(l)} \sim \mathcal{N}(0, 1)$.

The prediction for a test point is made by drawing $K$ samples and propagating them across the DGP network until the 
$L$-th layer using \eqref{Eq_montcarlo}. We denote $f_{*}^{(L)}$ as the latent function value at 
a new test location $\mathbf{x}_{*}$ in the last layer, $L$. The approximate predictive distribution for 
$f_{*}^{(L)}$ is
\begin{equation}
    q(\mathbf{f}_{*}^{(L)}) \approx \frac{1}{K}\sum_{k=1}^{K} 
	q(\mathbf{f}_{*}^{(L)}|\mathbf{r}^{(L)}, \mathbf{S}^{(L)}, \hat{\mathbf{f}}_{*}^{(L-1),k}, \mathbf{Z}^{(L)} )
\label{eq:dgp_pred}
\end{equation}
where $\hat{\mathbf{f}}_{*}^{(L-1),k}$ denotes the $k$-th sample from layer $L-1$.

Importantly, the formulation in \eqref{Eq_elbo_dgp} also allows for mini-batch sampling to train the model, 
which enables scaling to very large datasets. The predictive distribution for $y_* \in \mathds{R}$ 
can be easily obtained in the case of a Gaussian likelihood. One only has to incorporate the variance of the
additive Gaussian noise in \eqref{eq:dgp_pred}.

\subsection{Missing Gaussian Process}\label{MGP}

In this subsection, we introduce our method, missing Gaussian process (MGP), to impute missing values.
In many practical problems, the imputed value from one dimension highly depends on the values from 
all other dimensions as it has been observed in \cite{royston2011multiple}. Specifically, chained equation 
algorithms have been very successful in imputing missing values, as explained in the related work section. 
Here, we propose a new version of these algorithms based on chained Gaussian processes. 
The idea of MGP is inspired from DGPs \cite{salimbeni2017doubly} 
and recurrent Gaussian processes \cite{mattos2016recurrent}, where output of each GP depends on a previous GP.

Let us denote the $D$-dimensional input matrix of observed data with 
$\hat{\mathbf{X}} = (\hat{\mathbf{x}}_1,...,\hat{\mathbf{x}}_N)^{T}$, where 
$\hat{\mathbf{x}}_i$ is the $i$-th sample that randomly has some missing values, 
which have been initially imputed with the mean of the observed values at each dimension. 
The total attributes containing missing values are denoted by $D_m$. We sort 
these variables according to their standard deviations (before standardizing the data), 
from lowest to highest. For example, the attribute with the smallest standard deviation becomes the 
first attribute, followed by the attribute with the second smallest standard deviation, etc. The 
attributes without missing values are left as the last ones in the ordering. The ordering of
these attributes is not important.

\begin{figure}
  \includegraphics[width=\linewidth]{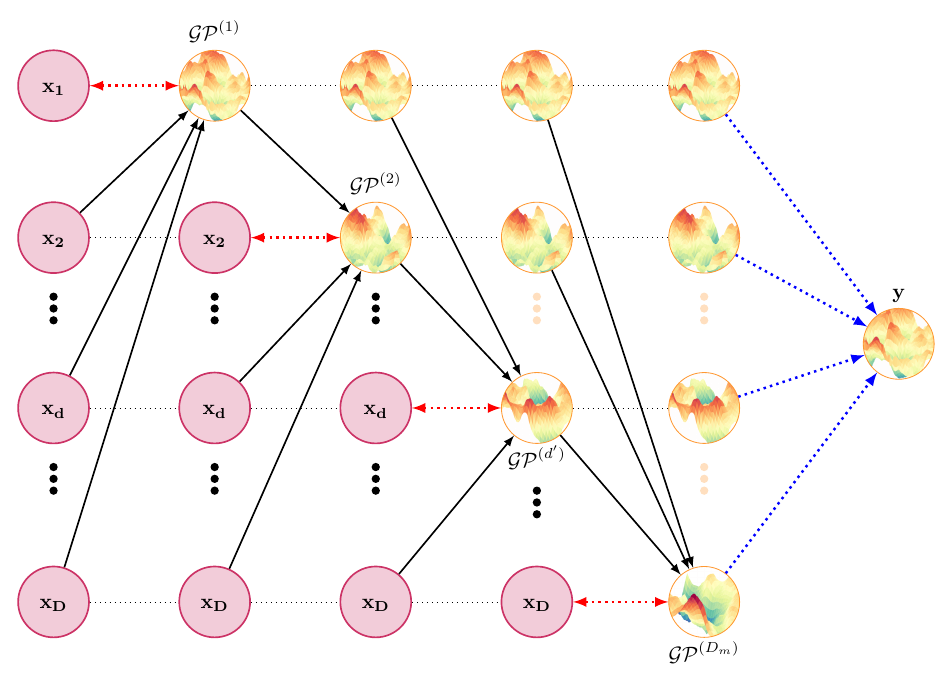}
  \caption{Missing GP architecture. It consists of $D_m$ GPs, where $D_m$ is the number of missing dimensions in the data. Each GP is used to impute the missing value in the dimension $d'$. The input of $GP^{(d')}$ are all other dimensions excluding $d$. The missing values of dimension $1,..., d-1$ are imputed using the predictions of the corresponding $GPs$. Black arrows show connections. Red arrows demonstrate log-likelihood computation. Dahsed black lines are copying the information.}
  \label{fig_mgp}
\end{figure}

Our method works as follows. First, it uses a GP to predict the missing values corresponding to 
the first attribute (after the ordering of the attributes) in terms of all the other attributes.
After this step, all missing values corresponding to the first attribute are replaced with the 
corresponding GP predictions, which are random variables determined by the GP 
predictive distribution. The first attribute, alongside with the GP predictions for missing values
and all the other attributes, are fed to a second GP to predict the missing values corresponding 
to the second attribute (after the ordering of the attributes). Therefore, some inputs to this
second GP are random variables, \emph{i.e.}, those corresponding to the missing values of the 
first attribute. Next, the first and two attributes, where missing values are replaced by the corresponding GP
predictive distribution, are fed, alongside with the remaining attributes, to a third GP to predict the
missing values corresponding to the third attribute (after the ordering of the attributes). The process will
be iterated until all dimensions with missing values have an associated GP that predicts their values.
The observed input attributes alongside with the GP predictions for missing values are then feed to a 
final GP that predicts the target variable. Figure \ref{fig_mgp} shows the architecture described.
The resulting method can hence be understood as a particular case of a DGP in which some GPs 
in some layers predict the missing values in the observed data.

Let $D_m$ be the total number of dimensions with associated missing values.
Let $\mathbf{f}^{(l)}\in \mathds{R}^N$ be the process values at layer $l$ for the training data.
Similarly, let $\mathbf{u}^{(l)} \in \mathds{R}^M$ be the process values at layer $l$ for the inducing points.
We denote the input of the GP at layer $l$ with $\tilde{\mathbf{X}}^{(l-1)}$. 
This is a $N \times D$ matrix equal to $\hat{\mathbf{X}}$, but where missing values
corresponding to dimensions $1$ to $l-1$ are replaced by the corresponding GP predictions
of the previous layer. Therefore, $\tilde{\mathbf{X}}^{(l-1)}$ can be computed in terms of
$\hat{\mathbf{X}}$ and $\{\mathbf{f}^{(l'-1)}\}_{l'=1}^{l-1}\}$, the predictions of the 
previous GPs in the sequence. For the first GP, we simply define 
$\tilde{\mathbf{X}}^{(0)}=\hat{\mathbf{X}}$. In the last GP, the input is $\tilde{\mathbf{X}}^{(D_m)}$.
The joint distribution of all the observed variables in our model is 
\begin{align}
\label{Eq_mgp_p}
p(\mathbf{y}, \{\mathbf{f}^{(l)},\mathbf{u}^{(l)}\}_{l=1}^{D_m+1}) & = 
\prod_{l=1}^{D_m+1} p(\mathbf{f}^{(l)}|\mathbf{u}^{(l)}, \tilde{\mathbf{X}}^{(l-1)}, \mathbf{Z}^{(l)}) 
	p(\mathbf{u}^{(l)}| \mathbf{Z}^{(l)}) \times \nonumber \\
	& \quad \times \prod_{i=1}^{N} p(y_{i}| f_i^{(D_m+1)},\tilde{\mathbf{x}}_i^{(D_m)})
	\times  \prod_{i=1}^N  \prod_{l \notin \mathcal{M}_i} p(x_{i,l}|f_i^{(l)})
\end{align}
where $\mathcal{M}_i$ is the set of attributes with missing values associated to instance $\mathbf{x}_i$
and $p(x_{i,l}|f_i^{(l)})=\mathcal{N}(x_{i,l}|f^{(l)}(\tilde{\mathbf{x}}_i^{(l-1)}),\sigma^2_{l})$.
That is, we assume a Gaussian likelihood for predicting the corresponding observed values of an 
attribute with missing values. This is just a particular case of the DGP model described in the
previous section, but with extra likelihood factors. 

Similar to \eqref{Eq_dgp_q}, the variational distribution $q$ is defined
\begin{align}
q(\{\mathbf{f}^{(l)}, \mathbf{u}^{(l)}\}_{l=1}^{D_m+1}) &=  \prod_{l=1}^{D_m +1} 
	p(\mathbf{f}^{(l)}|\mathbf{u}^{(l)}, \tilde{\mathbf{X}}^{(l-1)}, \mathbf{Z}^{(l)}) q(\mathbf{u}^{(l)}) \,,
\end{align}
where we can again marginalize out all $\mathbf{u}^{(l)}$ in closed form to obtain
\begin{equation}
\label{Eq_mgp_q}
\begin{split}
q(\{\mathbf{f}^{(l)}\}_{l=1}^{D_m+1}) &= \prod_{l=1}^{D_m+1} q(\mathbf{f}^{(l)}|\mathbf{r}^{(l)}, \mathbf{S}^{(l)}; 
	\tilde{\mathbf{X}}^{(l-1)}, \mathbf{Z}^{(l-1)}) \\&=
\prod_{l=1}^{D_m+1} \mathcal{N}(\mathbf{f}^{(l)}|\bm{\mu}^{(l)}, \mathbf{\Sigma}^{(l)}).
\end{split}
\end{equation}
Where $\mathbf{\mu'}^{(l)}$ and $\mathbf{\Sigma'}^{(l)}$ are computed as in \eqref{Eq.SGPvar0} and \eqref{Eq.SGPvar}. 
Then, the variational ELBO of MGP is 
\begin{align}
	\mathcal{L}_{MGP} &= \sum_{i=1}^N \mathds{E}_q[\log p(y_i|\tilde{\mathbf{x}}_i^{(D_m)})] 
	+ \sum_{i=1}^{N} \sum_{l \notin \mathcal{M}_i}
	\mathds{E}_q[\log p(x_{i,l}|f_i^{(l)})] \nonumber\\
	\quad & - \sum_{l=1}^{D_m+1} \text{KL}[q(\mathbf{u}^{(l)})|p(\mathbf{u}^{(l)}| \mathbf{Z}^{(l)})]\,,
\label{Eq_elbo_mgp}
\end{align}
where the required expectations can be approximated via Mote Carlo simply by 
propagating samples through the GP network displayed in Figure \ref{fig_mgp}, as in the case of a DGP.
Importantly, our formulation optimizes all hyper-parameters 
and variatiaional parameters at the same time  by maximizing $\mathcal{L}_{MGP}$.
Algorithm \ref{alg:ALG1} shows the training details of MGP. This algorithm uses a mini-batch 
to obtain a noisy estimate of (\ref{Eq_elbo_mgp}) and its gradient, which is then used to update the parameters of 
each $q(\mathbf{u}^{(l)})$ and the hyper-parameters. The data-dependent term of (\ref{Eq_elbo_mgp}) is corrected
to account for the fact that it is estimated using a single mini-batch.

When making a prediction for a new data instance $\mathbf{x}_\star$, one can propagate 
$K$ samples through the GP network. This results in a Gaussian mixture to predict the latent 
function at layer $D_m+1$. That is,
\begin{equation}
    q(f_\star^{(D_m+1)}) \approx \frac{1}{K}\sum_{k=1}^{K} q(f_\star^{(D_m+1)}|\mathbf{r}^{(D_m+1)}, 
	\mathbf{S}^{(D_m+1)}, \tilde{\mathbf{x}}_i^{(D_m),k}, \mathbf{Z}^{(D_m+1)} )\,,
\end{equation}
where $\tilde{\mathbf{x}}_i^{(D_m),k}$ is the $k$-th sample of $\tilde{\mathbf{x}}_i^{(D_m)}$, the
input to the last GP in the network. Again, when making predictions for the target variable, $y_\star$,
one simply has to add the variance of the additive Gaussian noise to each component of the 
previous Gaussian mixture.

	\begin{algorithm}[h!]
		\caption{Training algorithm for MGP}
		\label{alg:ALG1}
		\begin{algorithmic}
			\REQUIRE Training data $\mathcal{D}$ and $D_m$ attributes with missing values, 
			initial matrix of observed attributes $\hat{\mathbf{X}}$, and $M$ inducing points for each layer.
			\ENSURE Optimal parameters of the model
			\\
			Initialize model's hyper-parameters $\boldsymbol{\theta}$
			
			\WHILE{stopping criteria is False}
			\STATE Gather mini-batch $\mathbf{mb}$ of size $n$ from $\mathcal{D}$
			\FOR {($\mathbf{x_{i}}$) in $\mathbf{mb}$ }
			\STATE {Propagate samples and compute $\hat{f}_{i}^{(l)}$ for each layer $l$}
			\ENDFOR
			\STATE Estimate ELBO using \eqref{Eq_elbo_mgp} and the propagated samples $\hat{f}_{i}^{(l)}$: \\
			$\hat{\mathcal{L}}_{MGP}  \gets \frac{N}{n} \times \log_\text{lk} - \text{KL}$
			\STATE Update parameters of the model using the noisy gradient of $\hat{\mathcal{L}}_{MGP}$
			\ENDWHILE
		\end{algorithmic}
	\end{algorithm}

\section{Related Work}

A Gaussian mixture model (GMM) trained with the expectation-maximization algorithm 
has been proposed to impute missing values based on the acquired class 
\citep{schafer1997analysis, melchior2018filling}. Similarly, K-nearest neighbors (KNN) 
\citep{batista2002study} has also been proposed to impute missing values. 
This method does not rely on a predictive model for the missing variable. 
However, its estimation accuracy is strongly affected by the number of neighbors. 
Self-organizing maps (SOM) \citep{folguera2015self} has also been used for data 
correction and imputation for continuous and categorical data.
These techniques, \emph{i.e.}, GMM, KNN and SOM, do not require an iterative 
process to impute the missing values, unlike our method MGP.  
However, their performance is expected to be limited for the same reason.
Specifically, the iterative process that MGP follows can use partial imputed 
information for some missing values to predict other missing values.

Multiple imputations using chained equations (MICE) \cite{royston2011multiple} 
is another state-of-the-art algorithm that uses $S$ different imputation of missing 
values using $Z$ linear regression models. It considers variables with missing values 
as the dependent variable for each model. We compare results in our experiments with
this technique showing improved results. We believe this may be due to the extra 
flexibility of GPs for missing value imputation compared to the linear regression 
models described.

Recently, more sophisticated methods such 
as Auto-encoders (AE) \cite{lin2020missing,ning2017missing}, variational AE 
\cite{pereira2020vae}, and heterogeneous-incomplete VAE \cite{nazabal2020handling} have 
been proposed to impute missing values. In general, AE based methods use neural networks to 
impute missing values. Generative adversarial network (GAIN) for missing data imputation 
\cite{yoon2018gain} is another method based on neural 
networks. In GAIN, a generator neural network is 
used to generate new values for missing values. Similarly, a discriminator neural network is 
used for training the discriminator efficiently. We compare our method MGP with 
GAIN imputation showing improved results. We believe the reason for this is that 
GAIN is expected to perform well in big datasets. By contrast, a GP based approach
is expected to perform better in a small data regime. The reason for this is 
the robustness of the GP predictive distribution that incorporates
uncertainty estimation about the predictions made. 

There is a few studies on using GP based methods 
for imputing missing values. 
In \cite{owhadi2022computational}, the authors converted the missing data imputation problem into a computational graph completion problem.  Variables are denoted by the graph nodes and edges represent functions. The missing variables are replaced by the predictions of independent GPs that recover the unobserved variables in terms of the observed variables. This principle is similar to that of MGP. However, the way they operate is very different. They use Maximum a Posteriori (MAP) estimation to tune hyper-parameters of each GP instead of approximate marginal likelihood estimation. Moreover, they ignore the GP predictive variance and simply replace each missing value with the GP predictive mean \cite{williams2006gaussian}. Moreover, the covariance function hyper-parameters are assumed to be given beforehand. Finally, unlike MGP, each GP does not receive inputs from a previous GP and is isolated.
They do not consider the case of having several missing values associated with the same data instance, unlike MGP
In particular, \cite{fortuin2020gp} proposes a 
combination of GP and VAE for imputing missing values. According to our knowledge, there is no study on 
imputing missing values using deep GPs \cite{salimbeni2017doubly} nor SVGP \cite{hensman2015scalable}.
The proposed model from \cite{fortuin2020gp} is used GP in the latent space of VAE to model time series and impute missing values. The model is exclusively working on multi-variate time series data. Moreover, it has a fixed GP kernel which can not benefit from joint optimization.
In our work, we use a network of SVGPs that resembles a deep GP to impute missing 
values after mean pre-imputation of missing values. MGP learns from the observed value 
of each attribute and, similar to what happens in MICE, it uses previously 
imputed missing values for this task.

\section{Experimental Setup}\label{Exp}

We use five different data sets to evaluate the proposed method MGP. 
Table \ref{tab:data_info} describes the datasets. Four datasets are 
publicly available from UCI repository datasets \citep{Dua2019}.  
The last dataset called $TotalBrainVolume$ is a private dataset obtained from 
"Perinatal brain damage" group at Biomedical Research and Innovation Institute of 
Cádiz (INiBICA) Research Unit, Puerta del Mar University Hospital University of Cádiz, 
Spain. It is related to preterm infants and its different categorical and 
continuous attributes are the clinical information related to these infants. 
It initially contains $3.2$ percent missing values. All the datasets are 
standardized using Z-score transformation method. All categorical 
variables are converted to continuous variables using one-hot encoding strategy.
For each dataset, we generate five different splits, where $70$ percent of the data 
are used for training, and the rest, $30$ percent, are used for testing. 
Then, we randomly removed $10$, $20$, $30$, $40$ percents of the observed data 
in each dataset split to randomly introduce missing values. We report results for each 
different level of missing values. The performance of the proposed method, MGP, is compared to:
\begin{itemize}
    \item \textbf{Mean}: The mean value of each variable is computed and used to impute missing values.
    \item \textbf{Median}: The median value of each variable is used to impute missing values.
    \item \textbf{KNN}: A K-nearest neighbor is used to estimate and replace the missing values. 
	The number of neighbors is fixed to be $2$ in all the problems.
    \item \textbf{GAIN}: Generative adversarial network for missing data imputation \cite{yoon2018gain} 
	is also used to compute the missing values. The number of iterations are fixed to $20,000$. 
	The $\alpha$ value is set to be $10$, as recommended, and all the other specifications are 
	similar to what is suggested in \cite{yoon2018gain}. We observed that GAIN suffers from over-fitting and often 
	does not perform well on previously unseen data.  
    \item \textbf{MICE}: Multiple imputation using chained equations \cite{royston2011multiple} is another 
	state-of-the-are algorithm that has been used in this experiment. Linear regression is used 
	to estimate each missing value and the number of repetitions used is $10$.
    \item \textbf{SVGP}: Sparse variational Gaussian process \cite{hensman2015scalable}, as described 
	in Section \ref{SVGP}. Missing values are estimated using mean imputation. The number of inducing points 
	and the number of training iterations are fixed to be $100$ and $10,000$, respectively. 
    \item \textbf{DGP}: Five layered deep Gaussian process, as described in \cite{salimbeni2017doubly}, 
	and in Section \ref{DGP}. Again, we use mean imputation to estimate missing values. The 
	specifications are similar to $SVGP$. 
    \item \textbf{MGP}: Our proposed method. It is also trained for a total of $10,000$ iterations, except 
	for the TotalBrainVolume and Parkinson datasets where $2000$ iterations are used for training. 
\end{itemize}

The mini-batch size for all GP based algorithms is $100$. All GP based methods and GAIN are optimized 
using Adam \citep{kingma2014adam} and a learning rate equal to $0.01$. We use $20$ samples when training and 
testing in all GP based methods. All the experiments have been executing using two RTX A5000 GPUs (24 Gb), 
available at INiBICA.

	\begin{table}[H]
		\caption{Characteristics of the datasets.}
		\label{tab:data_info}
		\centering
		\begin{tabular}{l@{\hspace{0.2cm}}|l@{\hspace{0.2cm}}l@{\hspace{0.2cm}}}
			\bfseries Dataset & \bfseries N & \bfseries d \\
			\hline
			Protein & 45,730 & 10 \\
			KeggD & 53,414 & 23 \\
			KeggUD & 65,554 & 28 \\
			Parkinson & 1,040 & 24 \\
			TotalBrainVolume & 867 & 31 \\
			\hline
		\end{tabular}
	\end{table}

Although most of the methods described can be used to predict a target variable $y$ associated to
each dataset,  in our experiments we focus exclusively on missing value imputation. 
That is, we try to predict all missing values present in the data and do not consider a target variable $y$ to
be predicted. That is straight-forward to do in our proposed method, MGP, and other approaches we compare with. 
In DGP and SVGP (SVGP is just a DGP with one layer) we simply have at the output layer $D$ different GPs, one for 
each attribute with missing values. We then have a likelihood factor for each observed attribute.

We compare all methods in terms of the root mean squared error of missing value imputation in the test set. 
Namely,
\begin{equation}
    RMSE = \sqrt{\frac{1}{D} \sum_{d=1}^{D}  \frac{1}{N} \sum_{i=1}^{N} (X_i^d-X_i^{'d})^{2} }
\end{equation}
where $X_i^D$ is the $i$th true missing value and $X_i^{'d}$ is $i$th estimated value at dimension $D$.
In the GP based methods we use the mean of the predictive distribution as the model's prediction.

In all these experiments, we focus on regression inside each layer of MGP. 
However, one can use classification GPs, besides regression, whenever the output 
is binary, as in \citep{salimbeni2017doubly}. This also happens in the case of SVGP and DGP.

\section{Results and Discussion}\label{Results_discussion}

\begin{figure}
\centering
  \includegraphics[width=1.1\linewidth]{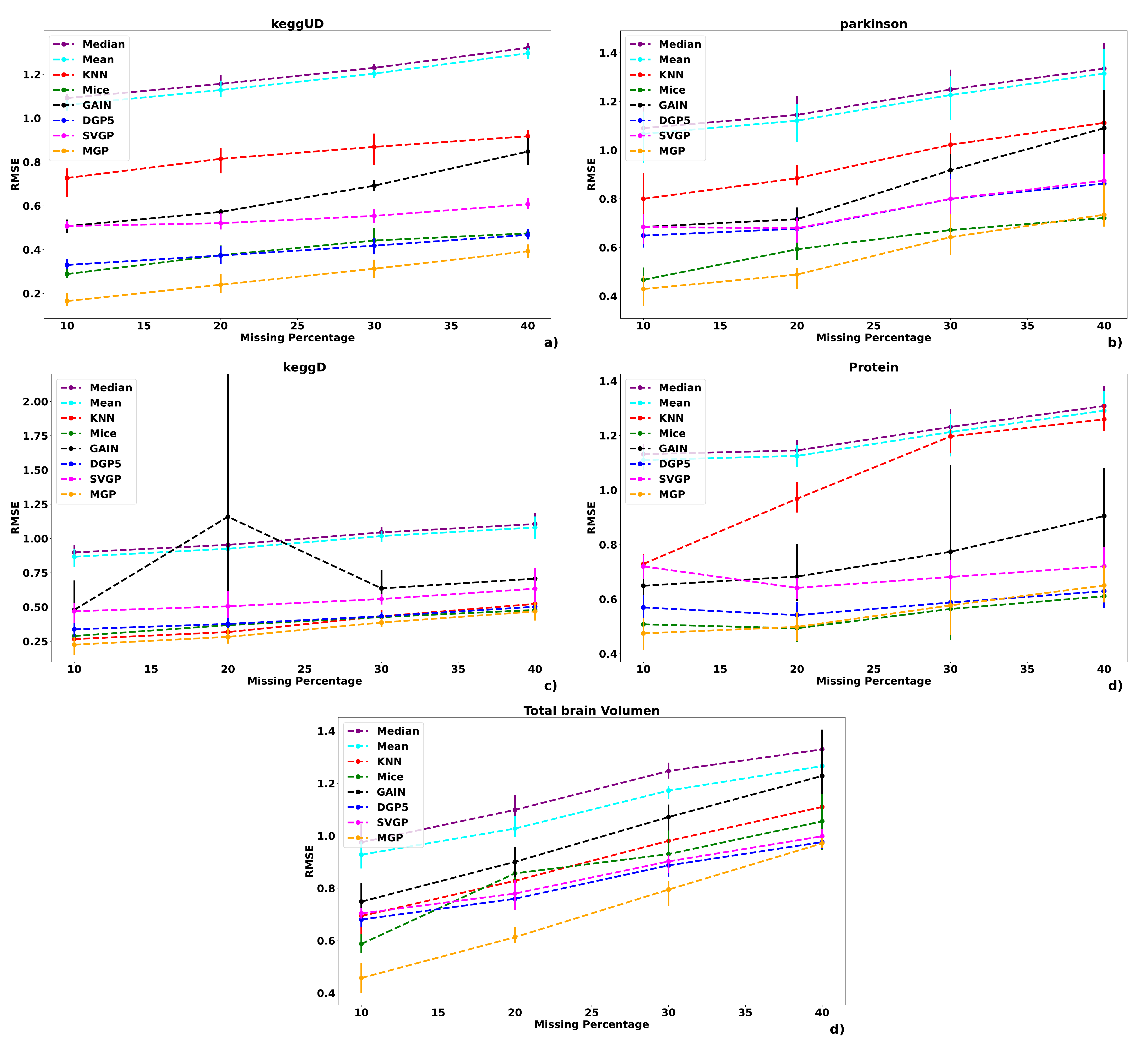}
  \caption{Average RMSE values obtained by the used methods in this study at various missing rates a) KeggUD, b) Parkinson, c) KeggD, d) Protein and e) Total brain volume. The error bar shows the minimum and maximum RMSE values obtained by each algorithm.}
  \label{fig_comparison}
\end{figure}

Tables \ref{tab:dataset_rmse_10} to \ref{tab:dataset_rmse_40} show the RMSE for each 
method after randomly removing $10\%$, $20\%$, $30\%$ and $40\%$ of the values from 
the data, respectively. We observe that the proposed algorithm MGP, most of the times, 
has a better performance than the other methods on each dataset. Figure \ref{fig_comparison} 
shows similar results graphically for each dataset and each level of missing values.  

In general, mean and median imputation based methods are the worst methods in all cases. 
KNN on KeggD, and TotalBrainVolume datasets has a comparable accuracy to $GP$ based methods, 
while in KeggUD, Parkinson and Protein its performance is worse.
MICE imputation is close to MGP in KeggD and Parkinson. 
GAIN method's performance is between that of SVGP and the mean imputation method.
SVGP and DGP perform similarly to each other on small dataset, \emph{i.e.}, 
Parkinson and TotalBrainVolume. In the other datasets DGP is better than $SVGP$.
MGP has very good accuracy when the level of missing values is low and, as this level
increases, its performance becomes closer to its GP-based variants and MICE.

\begin{figure}
\centering
  \includegraphics[width=0.7\linewidth]{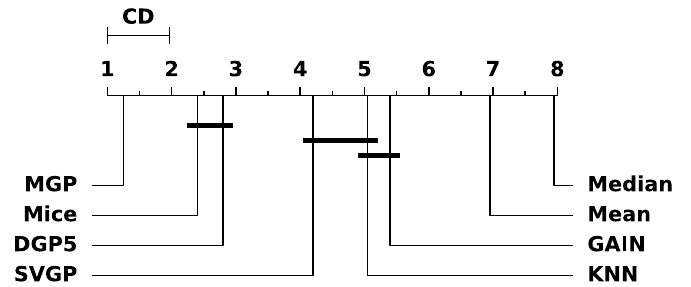}
  \caption{Average rank of each method alongside with the corresponding critical distance on all datasets and splits when considering all levels of missing values $10\%$, $20\%$, $30\%$ and $40\%$.}
  \label{fig_cdf}
\end{figure}

To get overall results, 
we computed the average rank of each method across all datasets splits and levels 
of missing values. In particular, if a method obtains the best performance for a dataset
split and level of noise, it gets rank 1. If it obtains the second best performance, it gets
rank 2, etc. Figure \ref{fig_cdf} shows the average rank of each method.
Following \cite{demvsar2006statistical}, we carried out a Nemenyi pot-hoc test to
look for statistical differences between average ranks.
If the average ranks of two methods are far apart one from another
by a distance bigger than the critical distance (CD) shown in Figure \ref{fig_cdf},
the differences are statistically significant. The critical distance is computed 
in terms of the number of methods compared, $8$, and the number of datasets and splits 
considered. Namely, $5 \times 5 = 25$. The reason for this is that the missing values are different in each 
dataset split.  We observe that mean and median imputations are the worst methods, 
overall. According to Figure \ref{fig_cdf}, MGP is the best performing method overall, followed by 
DGP and MICE, which have similar overall performance. SVGP, KNN, and GAIN perform similarly,
and the same happens for mean and median imputation.

\begin{figure}
\centering
  \includegraphics[width=0.7\linewidth]{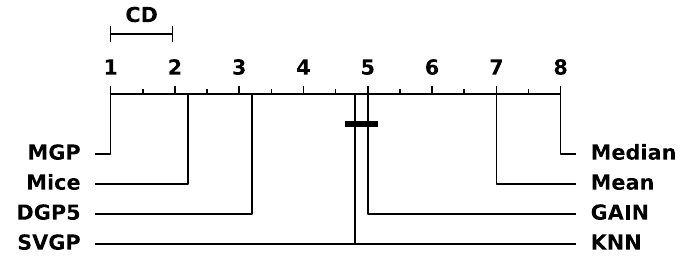}
  \caption{Average rank of each method alongside with the corresponding critical distance on all datasets and splits when considering the level of missing values $10\%$.}
  \label{fig_cdf_10}
\end{figure}

\begin{figure}
\centering
  \includegraphics[width=0.7\linewidth]{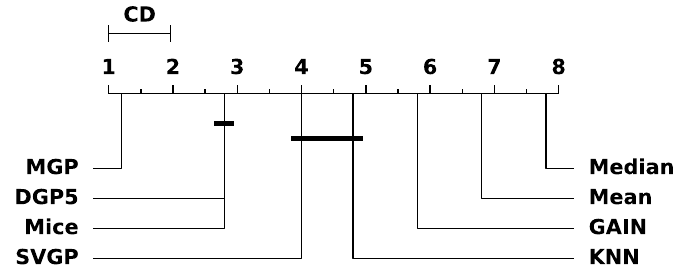}
  \caption{Average rank of each method alongside with the corresponding critical distance on all datasets and splits when considering the level of missing values $20\%$.}
  \label{fig_cdf_20}
\end{figure}

\begin{figure}
\centering
  \includegraphics[width=0.7\linewidth]{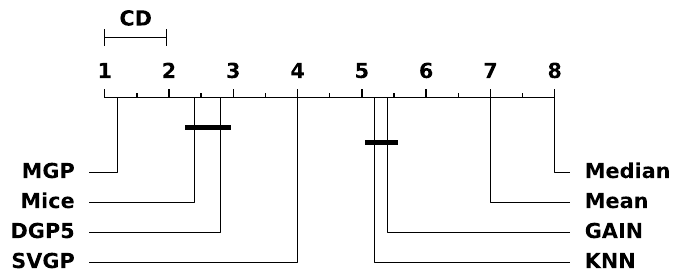}
  \caption{Average rank of each method alongside with the corresponding critical distance on all datasets and splits when considering the level of missing values $30\%$.}
  \label{fig_cdf_30}
\end{figure}

\begin{figure}
\centering
  \includegraphics[width=0.7\linewidth]{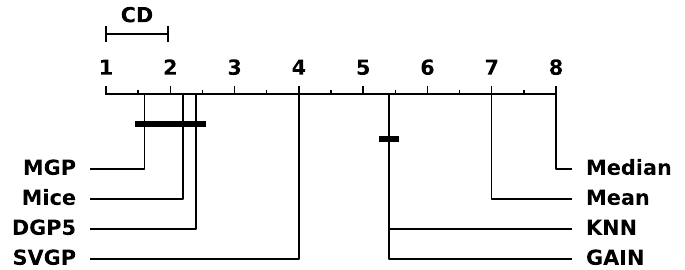}
  \caption{Average rank of each method alongside with the corresponding critical distance on all datasets and splits when considering the level of missing values $40\%$.}
  \label{fig_cdf_40}
\end{figure}

Figure \ref{fig_cdf_10} to \ref{fig_cdf_40} show similar results for each different level of missing values.
Namely, $10\%$, $20\%$, $30\%$ and $40\%$, respectively. We observe that in general the results are
similar to those of Figure \ref{fig_cdf} and  MGP is the best method overall. However, when the level
of missing values increases to $40\%$ the differences between MGP, MICE and DGP become smaller.

Our MGP implementation is coded using PyTorch and is publicly available\footnote{\url{https://github.com/BahramJafrasteh/MissingGPs}}.

	\begin{table}[H]
		\caption{Average RMSE values for $10\%$ missing values. 
	The numbers in parentheses are standard errors. Best mean values are highlighted.}
		\label{tab:dataset_rmse_10}
		\centering
		\begin{tabular}[t]{l@{\hspace{0.1cm}}c@{\hspace{0.1cm}}c@{\hspace{0.1cm}}c@{\hspace{0.1cm}}c@{\hspace{0.1cm}}c@{\hspace{0.1cm}}c@{\hspace{0.1cm}}c@{\hspace{0.1cm}}c@{\hspace{0.1cm}}}
			\hline
			& Protein & KeggD & KeggUD & Parkinson & TBV\\
			\hline
			Median & 1.13(0.02) & 0.90(0.04) & 1.09(0.01) & 1.09(0.08) & 0.98(0.05) \\
			Mean & 1.11(0.02) & 0.87(0.04) & 1.06(0.01) & 1.07(0.08) & 0.93(0.04)\\
			KNN & 0.73(0.03) & 0.27(0.02) & 0.73(0.04) & 0.80(0.06) & 0.69(0.06)\\
			MICE & 0.51(0.03) & 0.29(0.03) & 0.29(0.01) & 0.47(0.03) & 0.59(0.02)\\
			GAIN & 0.65(0.05) & 0.48(0.11) & 0.51(0.02) & 0.68(0.04) & 0.75(0.04) \\
			DGP  & 0.57(0.03) & 0.34(0.04) & 0.33(0.01) & 0.65(0.03) & 0.68(0.02) \\
			SVGP & 0.72(0.03) & 0.47(0.04) & 0.51(0.01) & 0.68(0.04) & 0.70(0.01)\\
			MGP & \bfseries 0.47(0.04) & \bfseries 0.23(0.04) & \bfseries 0.17(0.02) & \bfseries 0.43(0.05) & \bfseries 0.46(0.05) \\
			\hline
		\end{tabular}
	\end{table}

	\begin{table}[H]
		\caption{Average RMSE values for $20\%$ missing values. 
		The numbers in parentheses are standard errors. Best mean values are highlighted.}
		\label{tab:dataset_rmse_20}
		\centering
		\begin{tabular}[t]{l@{\hspace{0.1cm}}c@{\hspace{0.1cm}}c@{\hspace{0.1cm}}c@{\hspace{0.1cm}}c@{\hspace{0.1cm}}c@{\hspace{0.1cm}}c@{\hspace{0.1cm}}c@{\hspace{0.1cm}}c@{\hspace{0.1cm}}}
			\toprule
			& Protein & KeggD & KeggUD & Parkinson & TBV\\
			\midrule
			Median & 1.15(0.03) & 0.95(0.04) & 1.16(0.03) & 1.14(0.06) & 1.10(0.04) \\
			Mean & 1.13(0.03) & 0.93(0.04) & 1.13(0.03) & 1.12(0.05) & 1.03(0.03)\\
			KNN & 0.97(0.04) & 0.32(0.02) & 0.81(0.04) & 0.88(0.03) & 0.83(0.03)\\
			MICE & \bfseries  0.49(0.03) & 0.37(0.01) & 0.37(0.02) & 0.59(0.03) & 0.86(0.03)\\
			GAIN & 0.68(0.08) & 1.16(1.37) & 0.57(0.01) & 0.72(0.04) & 0.90(0.04) \\
			DGP & 0.54(0.03) & 0.38(0.07) & 0.37(0.03) & 0.68(0.03) & 0.76(0.03)\\
			SVGP & 0.64(0.02) & 0.51(0.07) & 0.52(0.03) & 0.68(0.04)& 0.78(0.03)\\
			MGP & \bfseries 0.50(0.03) & \bfseries 0.28(0.04) & \bfseries 0.24(0.03) & \bfseries 0.48(0.03) & \bfseries 0.61(0.02) \\
			\bottomrule
		\end{tabular}
	\end{table}

	\begin{table}[H]
		\caption{Average RMSE values for $30\%$ missing values. 
		The numbers in parentheses are standard errors. 
		Best mean values are highlighted.}
		\label{tab:dataset_rmse_30}
		\centering
		\begin{tabular}[t]{l@{\hspace{0.1cm}}c@{\hspace{0.1cm}}c@{\hspace{0.1cm}}c@{\hspace{0.1cm}}c@{\hspace{0.1cm}}c@{\hspace{0.1cm}}c@{\hspace{0.1cm}}c@{\hspace{0.1cm}}c@{\hspace{0.1cm}}}
			\toprule
			& Protein & KeggD & KeggUD & Parkinson & TBV\\
			\midrule
			Median & 1.23(0.06) & 1.04(0.02) & 1.23(0.01) & 1.25(0.06) & 1.25(0.02) \\
			Mean & 1.21(0.06) & 1.02(0.02) & 1.20(0.01)& 1.23(0.06) & 1.17(0.02)\\
			KNN & 1.20(0.03) & 0.43(0.03) & 0.87(0.05) & 1.02(0.04) & 0.98(0.02)\\
			MICE & \bfseries0.56(0.06) & 0.43(0.01) & 0.44(0.03) & 0.67(0.06) & 0.93(0.08)\\
			GAIN & 0.77(0.18) & 0.64(0.07) & 0.69(0.01) & 0.92(0.05) & 1.07(0.04)\\
			DGP & 0.59(0.06) & 0.43(0.02) & 0.42(0.02) & 0.80(0.05) & 0.89(0.02)\\
			SVGP  & 0.68(0.06) & 0.56(0.02) & 0.55(0.02) & 0.80(0.05) & 0.90(0.02)\\
			MGP &  0.58(0.06) & \bfseries 0.39(0.02) & \bfseries 0.31(0.02) & \bfseries 0.64(0.06) & \bfseries 0.79(0.03)\\
			\bottomrule
		\end{tabular}
	\end{table}

	\begin{table}[H]
		\caption{Average RMSE values for $40\%$ missing values. 
		The numbers in parentheses are standard errors. Best mean values are highlighted.}
		\label{tab:dataset_rmse_40}
		\centering
		\begin{tabular}[t]{l@{\hspace{0.1cm}}c@{\hspace{0.1cm}}c@{\hspace{0.1cm}}c@{\hspace{0.1cm}}c@{\hspace{0.1cm}}c@{\hspace{0.1cm}}}
			\toprule
			& Protein & KeggD & KeggUD & Parkinson & TBV\\
			\midrule
			Median & 1.31(0.05) & 1.11(0.05) & 1.32(0.01) & 1.34(0.06) & 1.33(0.02) \\
			Mean & 1.29(0.05) & 1.08(0.05) & 1.30(0.01)& 1.31(0.05) & 1.27(0.02)\\
			KNN & 1.26(0.04) & 0.52(0.03) & 0.92(0.03) & 1.11(0.05) & 1.11(0.01)\\
			MICE & \bfseries0.61(0.03)& \bfseries 0.48(0.03) & 0.47(0.01) & \bfseries0.72(0.03) & 1.06(0.12)\\
			GAIN & 0.90(0.13) & 0.71(0.06) & 0.85(0.04) & 1.09(0.10) & 1.23(0.09)\\
			DGP & 0.63(0.05) & 0.50(0.07) & 0.47(0.01) & 0.86(0.04) & 0.98(0.02) \\
			SVGP & 0.72(0.05) & 0.63(0.09) & 0.61(0.02) & 0.87(0.05) & 1.00(0.02)\\
			MGP &  0.65(0.05) & \bfseries 0.47(0.04) & \bfseries 0.39(0.02) & \bfseries 0.73(0.06) & \bfseries 0.97(0.01)\\
			\bottomrule
		\end{tabular}
	\end{table}

\section{Conclusions}\label{sec:conclusion}

We have presented a novel hierarchical composition of sparse variational GPs 
to impute missing values, inspired by deep GPs and recurrent GPs. The proposed method, 
called missing GP (MGP), has been evaluated on four UCI benchmark data sets and on one 
real-life private medical dataset, where $10$, $20$, $30$, and $40$ percent of the data attributes
contain missing values. 

In our experiments, we observed a statistically significant better performance of MGP 
than other state-of-the-art methods for missing value imputation. Namely, KNN, MICE, GAIN, 
and mean and median imputation.  We also 
observed that MGP provides better results than other methods, \emph{i.e.},
deep GPs and sparse variational GPs (SVGP). In particular, when the fraction 
of missing values is not very high. By contrast, when this fraction is high, we believe 
that there is not enough data to train the sparse GPs inside MGP and the gains obtained 
are better, but not as significant.

In our work, we used only regression GPs inside MGP. However, it is also 
possible to use a combination of classification or regression layers to impute 
missing values with binary attributes. This will make approximate inference
more challenging since the Bernoulli distribution is not re-parametrizable.
However, it may be possible to leave the binary attributes as the last ones in the
hierarchical structure of MGP so that their output is not used for the
imputation of other variables.

\subsection*{Acknowledgement}
This study was funded  by the  Cadiz integrated territorial initiative for biomedical research, European Regional Development Fund (ERDF) 2014–2020. Andalusian Ministry of Health and Families, Spain. Registration number: ITI-0019-2019.
DHL acknowledges financial support from Spanish Plan Nacional I+D+i, grant PID2019-106827GB-I00/AEI/10.13039/501100011033.


\bibliography{mybibfile}

\begin{thebibliography}{10}
\expandafter\ifx\csname url\endcsname\relax
  \def\url#1{\texttt{#1}}\fi
\expandafter\ifx\csname urlprefix\endcsname\relax\def\urlprefix{URL }\fi
\expandafter\ifx\csname href\endcsname\relax
  \def\href#1#2{#2} \def\path#1{#1}\fi

\bibitem{Missingdata01}
R.~J. Little, D.~B. Rubin, Statistical analysis with missing data, Vol. 793,
  John Wiley \& Sons, 2019.

\bibitem{bishop2006}
C.~M. Bishop, Pattern Recognition and Machine Learning (Information Science and
  Statistics), Springer, 2006.

\bibitem{beaulieu2017missing}
B.~K. Beaulieu-Jones, J.~H. Moore, P.~R. O.-A. A. C.~T. CONSORTIUM, Missing
  data imputation in the electronic health record using deeply learned
  autoencoders, in: Pacific symposium on biocomputing 2017, World Scientific,
  2017, pp. 207--218.

\bibitem{ryu2020denoising}
S.~Ryu, M.~Kim, H.~Kim, Denoising autoencoder-based missing value imputation
  for smart meters, IEEE Access 8 (2020) 40656--40666.

\bibitem{villacampa2020multi}
C.~Villacampa-Calvo, B.~Zaldivar, E.~C. Garrido-Merch{\'a}n,
  D.~Hern{\'a}ndez-Lobato, Multi-class gaussian process classification with
  noisy inputs, arXiv preprint arXiv:2001.10523 (2020).

\bibitem{williams2006gaussian}
C.~K. Williams, C.~E. Rasmussen, Gaussian processes for machine learning,
  Vol.~2, MIT press Cambridge, MA, 2006.

\bibitem{titsias2009variational}
M.~Titsias, Variational learning of inducing variables in sparse gaussian
  processes, in: Artificial intelligence and statistics, PMLR, 2009, pp.
  567--574.

\bibitem{snelson2006sparse}
E.~Snelson, Z.~Ghahramani, Sparse gaussian processes using pseudo-inputs,
  Advances in neural information processing systems 18 (2006) 1257.

\bibitem{hensman2015mcmc}
J.~Hensman, A.~G. d.~G. Matthews, M.~Filippone, Z.~Ghahramani, Mcmc for
  variationally sparse gaussian processes, arXiv preprint arXiv:1506.04000
  (2015).

\bibitem{villacampa2017scalable}
C.~Villacampa-Calvo, D.~Hern{\'a}ndez-Lobato, Scalable multi-class gaussian
  process classification using expectation propagation, in: International
  Conference on Machine Learning, PMLR, 2017, pp. 3550--3559.

\bibitem{hensman2015scalable}
J.~Hensman, A.~Matthews, Z.~Ghahramani, Scalable variational gaussian process
  classification, in: Artificial Intelligence and Statistics, PMLR, 2015, pp.
  351--360.

\bibitem{damianou2013deep}
A.~Damianou, N.~D. Lawrence, Deep gaussian processes, in: Artificial
  intelligence and statistics, PMLR, 2013, pp. 207--215.

\bibitem{bui2016deep}
T.~Bui, D.~Hern{\'a}ndez-Lobato, J.~Hernandez-Lobato, Y.~Li, R.~Turner, Deep
  gaussian processes for regression using approximate expectation propagation,
  in: International conference on machine learning, PMLR, 2016, pp. 1472--1481.

\bibitem{salimbeni2017doubly}
H.~Salimbeni, M.~Deisenroth, Doubly stochastic variational inference for deep
  gaussian processes, arXiv preprint arXiv:1705.08933 (2017).

\bibitem{mattos2016recurrent}
C.~L.~C. Mattos, Z.~Dai, A.~Damianou, J.~Forth, G.~A. Barreto, N.~D. Lawrence,
  Recurrent gaussian processes (2016).

\bibitem{Dua2019}
D.~Dua, C.~Graff, \href{http://archive.ics.uci.edu/ml}{{UCI} machine learning
  repository} (2017).
\newline\urlprefix\url{http://archive.ics.uci.edu/ml}

\bibitem{rasmussen2005book}
C.~E. Rasmussen, C.~K.~I. Williams, {G}aussian Processes for Machine Learning
  (Adaptive Computation and Machine Learning), The MIT Press, 2006.

\bibitem{hensman2013gaussian}
J.~Hensman, N.~Fusi, N.~D. Lawrence, Gaussian processes for big data, arXiv
  preprint arXiv:1309.6835 (2013).

\bibitem{rezende2014stochastic}
D.~J. Rezende, S.~Mohamed, D.~Wierstra, Stochastic backpropagation and
  approximate inference in deep generative models, in: International conference
  on machine learning, PMLR, 2014, pp. 1278--1286.

\bibitem{kingma2015variational}
D.~P. Kingma, T.~Salimans, M.~Welling, Variational dropout and the local
  reparameterization trick, Advances in neural information processing systems
  28 (2015) 2575--2583.

\bibitem{royston2011multiple}
P.~Royston, I.~R. White, Multiple imputation by chained equations (mice):
  implementation in stata, Journal of statistical software 45 (2011) 1--20.

\bibitem{schafer1997analysis}
J.~L. Schafer, Analysis of incomplete multivariate data, CRC press, 1997.

\bibitem{melchior2018filling}
P.~Melchior, A.~D. Goulding, Filling the gaps: Gaussian mixture models from
  noisy, truncated or incomplete samples, Astronomy and computing 25 (2018)
  183--194.

\bibitem{batista2002study}
G.~E. Batista, M.~C. Monard, et~al., A study of k-nearest neighbour as an
  imputation method., His 87~(251-260) (2002) 48.

\bibitem{folguera2015self}
L.~Folguera, J.~Zupan, D.~Cicerone, J.~F. Magallanes, Self-organizing maps for
  imputation of missing data in incomplete data matrices, Chemometrics and
  Intelligent Laboratory Systems 143 (2015) 146--151.

\bibitem{lin2020missing}
W.-C. Lin, C.-F. Tsai, Missing value imputation: a review and analysis of the
  literature (2006--2017), Artificial Intelligence Review 53~(2) (2020)
  1487--1509.

\bibitem{ning2017missing}
X.~Ning, Y.~Xu, X.~Gao, Y.~Li, Missing data of quality inspection imputation
  algorithm base on stacked denoising auto-encoder, in: 2017 IEEE 2nd
  International Conference on Big Data Analysis (ICBDA), IEEE, 2017, pp.
  84--88.

\bibitem{pereira2020vae}
R.~C. Pereira, P.~H. Abreu, P.~P. Rodrigues, Vae-bridge: Variational
  autoencoder filter for bayesian ridge imputation of missing data, in: 2020
  International Joint Conference on Neural Networks (IJCNN), IEEE, 2020, pp.
  1--7.

\bibitem{nazabal2020handling}
A.~Nazabal, P.~M. Olmos, Z.~Ghahramani, I.~Valera, Handling incomplete
  heterogeneous data using vaes, Pattern Recognition 107 (2020) 107501.

\bibitem{yoon2018gain}
J.~Yoon, J.~Jordon, M.~Schaar, Gain: Missing data imputation using generative
  adversarial nets, in: International conference on machine learning, PMLR,
  2018, pp. 5689--5698.

\bibitem{owhadi2022computational}
H.~Owhadi, Computational graph completion, Research in the Mathematical
  Sciences 9~(2) (2022) 1--33.

\bibitem{fortuin2020gp}
V.~Fortuin, D.~Baranchuk, G.~R{\"a}tsch, S.~Mandt, Gp-vae: Deep probabilistic
  time series imputation, in: International conference on artificial
  intelligence and statistics, PMLR, 2020, pp. 1651--1661.

\bibitem{kingma2014adam}
D.~P. Kingma, J.~Ba, Adam: A method for stochastic optimization, arXiv preprint
  arXiv:1412.6980 (2014).

\bibitem{demvsar2006statistical}
J.~Dem{\v{s}}ar, Statistical comparisons of classifiers over multiple data
  sets, The Journal of Machine Learning Research 7 (2006) 1--30.

\end{thebibliography}
\bibliographystyle{elsarticle-num}

\end{document}